\documentclass[11pt]{article}
\PassOptionsToPackage{table}{xcolor}
\usepackage[preprint]{acl}
\usepackage{times}
\usepackage{latexsym}
\usepackage[T1]{fontenc}
\usepackage[utf8]{inputenc}
\usepackage{microtype}
\usepackage{inconsolata}
\usepackage{amsmath}
\usepackage{enumitem}
\usepackage{amssymb}
\usepackage{mathtools}
\usepackage{amsthm}
\usepackage{algorithm}
\usepackage{algorithmic}
\usepackage{placeins}

\usepackage[most]{tcolorbox}

\definecolor{lightgray}{gray}{0.92}
\definecolor{citecolor}{HTML}{0071BC}

\newcommand{\g}[1]{\textcolor{gray}{#1}}

\usepackage{pifont}
\usepackage{listings}
\tcbuselibrary{listings} 
\newtcblisting{promptbox}[1]{
    enhanced,
    breakable,             
    title={#1},            
    colback=gray!5,        
    colframe=gray!75,      
    coltitle=white,
    fonttitle=\bfseries,
    listing only,          
    sharp corners,         
    boxrule=0.5pt,
    listing options={
        basicstyle=\ttfamily\small,
        breaklines=true,   
        breakatwhitespace=true,
        columns=fullflexible,
        keepspaces=true,
        showstringspaces=false
    }
}

\newcommand{\our}{PaceReasoner}
\newcommand{\LONG}{LONG}
\newcommand{\SHORT}{SHORT}
\usepackage{graphicx}
\usepackage{subcaption}
\usepackage{booktabs} 
\usepackage{multirow}

\usepackage{amsfonts}
\usepackage{nicefrac}

\title{PaceReasoner: Label-Free Learning to Answer Briefly from Long Reasoning}
\author{%
  \parbox{\textwidth}{%
    \centering
    Changguo Jia$^{1}$\thanks{Equal contribution.}\quad
    Zhiyou Xiao$^{1}$\footnotemark[1]\quad
    Xinbei Ma$^{2}$\footnotemark[1]\quad
    Geng Chen$^{1}$\\[2pt]
    Weiming Zhang$^{3}$\quad
    Yiming Xu$^{1}$\quad
    Tianqi Zhao$^{4}$\quad
    Minghui Zhou$^{1}$\thanks{Corresponding author.}\\[6pt]
    {\normalfont
      $^{1}$Peking University\quad
      $^{2}$National University of Singapore\\[2pt]
      $^{3}$Shanghai Jiao Tong University\quad
      $^{4}$Zhongguancun Laboratory\\[6pt]
    }%
    {\normalfont\ttfamily
      jiachangguo@stu.pku.edu.cn\quad
      xiaozhiyou@stu.pku.edu.cn \\
      xinbei.ma@nus.edu.sg\quad
      zhmh@pku.edu.cn
    }%
  }%
}

\begin{document}
\maketitle

\begin{abstract}

Extended chain-of-thought reasoning improves problem-solving performance but increases inference cost and latency, motivating more accurate answers with shorter reasoning.
We introduce \our{}, a label-free test-time reinforcement learning framework that uses a model's LONG responses to improve its SHORT-mode accuracy. Both modes share a trainable policy, creating a need for reliable supervision, sufficient SHORT training samples, and continued LONG responses to guide learning. \our{} combines all-response and LONG-only answer consensus through progressive, confidence-weighted supervision. A behavior reward gradually shifts from encouraging SHORT exploration to rewarding agreement with LONG answers, while a ratio controller regulates the balance between modes.
Across three model backbones and four benchmarks, \our{} improves average SHORT accuracy over the baselines by 7.4--13.5 percentage points, with fewer than 25 reasoning tokens per response on average. Ablations support the key components, and cross-dataset results demonstrate transfer beyond the adaptation set.
\end{abstract}

\section{Introduction}
\label{sec:introduction}

Large reasoning models improve problem-solving performance
by devoting additional test-time computation to extended
chain-of-thought reasoning
\citep{deepseek2025r1,qwen3technicalreport}.
However, longer reasoning increases token usage and response
latency, while unnecessarily extended reasoning may yield
limited additional gains \citep{chen2025do,ma2025cot}.
For simpler questions or applications with latency or cost
constraints, these gains may not justify the additional
computation.
Existing approaches manage this trade-off by learning when
to invoke explicit reasoning \citep{thinkless,adaptthink,rpo},
compressing reasoning traces through self-distillation
\citep{crisp}, or controlling and allocating reasoning budgets
\citep{li-etal-2026-steering,lin2026plan}.
These approaches primarily optimize the reasoning cost rather than the problem-solving capability of a low-cost mode itself. Therefore, they motivate a complementary question: how can
we improve the model's problem-solving accuracy when it
operates in shorter reasoning mode?



Our starting point is the accuracy gap between LONG and SHORT
modes of the same model, where zero-shot LONG achieves higher average accuracy than SHORT
across all three evaluated backbones.
This gap suggests that the model's own LONG responses may
provide useful supervision for improving SHORT without
reference answers.
Motivated by this observation, we formulate the problem as \textbf{
label-free test-time adaptation}: given a set of test questions
without access to their reference answers, our goal is to
\textbf{improve the problem-solving accuracy of SHORT mode while
keeping its reasoning brief}.

We introduce \our{}, a test-time
reinforcement learning framework that jointly optimizes LONG
and SHORT within a shared policy, using LONG answers to guide
SHORT learning without reference answers.
This approach faces three challenges.
First, \textbf{supervision reliability}: LONG responses may
disagree or converge on an incorrect answer.
Second, \textbf{SHORT learning opportunities}: the initially
stronger LONG mode may dominate generation, limiting the
samples available for improving SHORT.
Third, \textbf{continued LONG supervision}: encouraging SHORT
too strongly may leave too few LONG responses to support
further learning.

\our{} addresses these challenges through coordinated answer
and behavior rewards.
To improve \textbf{supervision reliability}, progressive
supervision combines answer consensus across all responses
with consensus from LONG responses alone, while confidence
weighting reduces reliance on uncertain LONG consensus.
To provide \textbf{SHORT learning opportunities}, a behavior
reward initially encourages SHORT exploration and gradually
shifts toward rewarding agreement with LONG answers.
To preserve \textbf{continued LONG supervision}, a ratio
controller adjusts this reward according to the proportion
of SHORT responses in the batch, counteracting mode imbalance.
Finally, Group Relative Policy Optimization (GRPO) updates
the complete responses from both modes, including mode
selection and subsequent generation.

Our evaluation on three model backbones and four benchmarks
shows that \our{} improves average SHORT-mode accuracy over
TTRL by 7.4--13.5 percentage points, while using fewer than
25 reasoning tokens per response on average for each backbone.
Our further ablation studies support the contributions of progressive
supervision and behavior control, while cross-dataset
experiments show that SHORT-mode improvements transfer
beyond the adaptation dataset.

In summary, our contributions are threefold:
\begin{itemize}
\item We introduce \our, which improves short reasoning using answer supervision from the same model's LONG responses on unlabeled target questions.
\item We coordinate progressive consensus supervision, an exploration-to-agreement reward, and ratio control to address noisy supervision, insufficient SHORT coverage, and loss of LONG evidence during joint training.
\item Across three models and four benchmarks, \our{} improves average SHORT accuracy over baselines at fewer than 25 reasoning tokens on average.  Ablations support each key component, while cross-dataset results demonstrate transfer
beyond the adaptation set.
\end{itemize}

\section{Related Work}
\label{sec:related}

\subsection{Reasoning Compression and Distillation}

Length-constrained reinforcement learning and pruning improve the accuracy--cost trade-off \citep{l1,thinkprune}. Progressive CoT removal internalizes intermediate steps \citep{implicitcot}, while on-policy distillation learns from teacher distributions on student-generated trajectories \citep{gkd}. Teacher conditioning includes privileged solutions in OPSD \citep{opsd}, conciseness instructions in CRISP \citep{crisp}, and hidden reasoning prefixes in Masked Self-Distillation \citep{maskedselfdistillation}. Alternatively, LONG answers guide learning without distribution matching. PaceReasoner supervises SHORT using sequence-level rewards from LONG answer consensus.

\paragraph{Adaptive Reasoning.}
Qwen3 supports hybrid reasoning \citep{qwen3technicalreport}. Thinkless decouples mode-selection and response optimization, while AdaptThink balances mode exploration through importance sampling \citep{thinkless,adaptthink}. RPO optimizes only the root token \citep{rpo}. Greater SHORT usage need not improve capability. Mode allocation governs supervision: LONG dominance restricts SHORT updates, whereas SHORT dominance reduces LONG evidence. PaceReasoner coordinates mixed-mode sampling to sustain SHORT learning and LONG supervision.

\subsection{Label-Free Test-Time Adaptation}

Self-consistency aggregates answers at inference \citep{selfconsistency}, and TTRL converts majority agreement into reinforcement-learning rewards \citep{ttrl}; RENT and Intuitor use intrinsic confidence \citep{rent,intuitor}. U-OPSD distills disagreeing completions using consensus-supported pseudo-solutions \citep{uopsd}, whereas TTPO distills agreeing rollouts and selectively penalizes disagreeing ones \citep{ttpo}. Both improve non-thinking performance; TTPO studies thinking-to-non-thinking transfer. Joint adaptation couples supervision and mode allocation: updates change LONG evidence, while mode frequencies affect voting pools and training opportunities. All-response consensus offers broader coverage but risks domination by weaker SHORT predictions; LONG-only consensus isolates deliberative evidence but may be sparse or incorrect. PaceReasoner combines progressive, confidence-weighted dual-consensus rewards with mode regulation to sustain supervision during joint adaptation.

\section{Problem Definition and Preliminaries}
\label{sec:problem}
\subsection{Test-time Adaptation}
Test-time adaptation aims to update a policy at inference time without additional ground-truth supervision. Given an initial policy $\pi_{\theta_0}$ and a set of test questions
$\mathcal{D}=\{q_j\}_{j=1}^{M}$, the adaptation stage updates
the policy to $\pi_\theta$ using rewards derived from its own inference. After the learning stage, we evaluate the updated policy on the same questions $\mathcal{D}$, where the generates final answers are scored against the reference answer. 
In this study, our goal is to improve performance in the short
chain-of-thought setting, defined in the following section.


\subsection{Reasoning Mode}
Chain-of-thought largely facilitates the reasoning capabilities
of an LLM. Given a question $q$, an LLM with parameters
$\theta$ generates reasoning text $z$ followed by a final
answer $a$. We denote the resulting output by $o=(z,a)$:
\begin{equation}
o=(z,a)\sim\pi_\theta(\cdot\mid q).
\label{eq:resulting_output}
\end{equation}
To enable efficient reasoning, we use system prompts to elicit two reasoning modes, \LONG{} and \SHORT{}. 
The LLM is prompted either to choose between the two reasoning modes or to follow a specified mode, and to format its response accordingly.

(1) In \LONG{} mode, the response contains explicit reasoning
followed by the final answer:
\[
\texttt{<reason>}~z~\texttt{</reason>}
\texttt{<answer>}~a~\texttt{</answer>}.
\]
(2) In \SHORT{} mode, the response gives the final answer directly,
without an explicit reasoning trace:
\[
\texttt{<not></not>}
\texttt{<answer>}~a~\texttt{</answer>}.
\]
Thus, under the specified output format, $z=\varnothing$ in
SHORT mode. The full prompt used during adaptation is provided
in Appendix~\ref{app:training-prompt}.
Therefore, the reasoning policy is denoted as
\[
(z,a)\sim\pi_\theta(\cdot\mid q,m),
\]
where $L$ and $S$ denote \LONG{} and \SHORT{} and both
modes share the same policy parameters $\theta$.

We define the reasoning cost of a response as the number of
generated reasoning tokens before the final answer, excluding
answer tokens. Let $\ell(z)$ denote this count. For mode $m$,
the expected reasoning cost is
\begin{equation}
    C_m=\mathbb{E}_{(z,a)\sim\pi_\theta(\cdot\mid q,m)}
       [\ell(z)].
\label{eq:r_cost}
\end{equation}

\subsection{Group Relative Policy Optimization}

Group Relative Policy Optimization (GRPO) is a reinforcement
learning algorithm for LLM post-training that has been shown to be effective for
reasoning tasks with outcome-based reward
\citep{shao2024deepseekmath}, where only the answer $a$ is compared with a verifiable reference, without requiring supervision for the intermediate reasoning process.

Given a question $q$, GRPO samples a group of $K$ responses
$\{o_i\}_{i=1}^{K}$ from the policy. Each response receives a
scalar reward $r_i$, and its advantage is computed relative to
the other responses in the group, used to update the policy optimization,
\begin{equation}
\widehat{A}_i =
\frac{r_i-\operatorname{mean}_{j}(r_j)}
     {\operatorname{std}_{j}(r_j)+\epsilon}.
\label{eq:grpo_ad}
\end{equation}
In our test-time adaptation setting, the reward $r_i$, is derived from the policy's \LONG{} mode of reasoning, transforming self-generated signals to improve token-efficient reasoning of \SHORT{} mode.

\section{\our}
\label{sec:method}

\begin{figure*}[t]
\centering
\includegraphics[width=\textwidth]{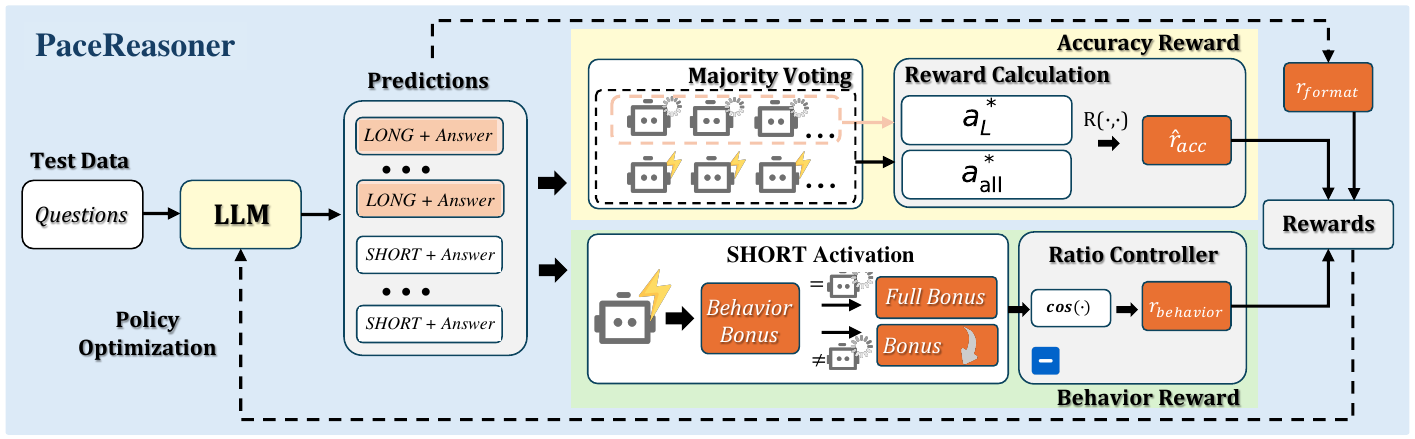}
\caption{Overview of \our. LONG and SHORT responses supply all-response and LONG-only answer consensus. The accuracy-reward block implements progressive supervision with the additional confidence weight in Eq.~\eqref{eq:accuracy_reward}. SHORT activation and ratio control construct the behavior reward; format, answer, and behavior rewards jointly drive policy optimization.}
\label{fig:main_pip}
\end{figure*}

\subsection{Method Overview}
Building on GRPO-based test-time adaptation, the primary idea of \our{} is to construct a reward that encourages \SHORT{} to generate the same answer as \LONG{}, enhancing short reasoning of the policy model.
We propose a novel reward that uses answer consensus from \LONG{}.
As is shown in Figure~\ref{fig:main_pip}, for each question, the current
policy performs $K$ times of rollout, where it chooses a reasoning mode and follows the format to generate $K$ responses. 
Since \LONG{} always reasons to superior answer accuracy than \SHORT{}, we use the predicted answer of \LONG{}'s consensus to align \SHORT{}.
Another key is the mode-choosing behavior:
because both modes are updated by the same policy, the training requires enough \SHORT{} for learning while preserving enough \LONG{} as a source of supervision. Therefore, we have to prevent the reasoning mode from collapsing into any single one.

The reward for $o_i=(m_i,z_i,a_i)$ is defined as
\begin{equation}
r_i=r_{\mathrm{fmt}}(o_i)+r_{\mathrm{acc}}(q,a_i,t)+r_{\mathrm{beh}}(o_i,t).
\label{eq:total_reward}
\end{equation}
The format reward is 1 for a valid  \LONG{} or \SHORT{} output and 0 otherwise. The answer reward supplies pseudo-supervision, while the behavior reward regulates reasoning mode exploration and prevalence.

\subsection{Answer Reward: Long-CoT Consensus}
\paragraph{Answer consensus.}
Let $\mathcal A_L(q)$ contain those generated in \LONG{} mode, whose majority voting yields a pseudo-label $a^*_L$. 
For response $o_i$, let $a_i$ denote its extracted final
answer, and let $a^*_L$ denote the majority-voted answer from
the corresponding \LONG{} answer pool. We define the answer-agreement
reward $R(a_i,a^*_L)$ to be 1 if the task-specific answer
checker judges $a_i$ equivalent to $a^*_L$, and 0 otherwise.
If either answer is missing, we set $R(a_i,a^*_L)=0$.

$\mathcal A_L(q)$ voting isolates evidence produced through explicit deliberation, yet remains vulnerable to limited coverage, disagreement, and shared errors. 
We therefore consider
All-response voting to provide feedback when \LONG{} samples are scarce, using all extractable answers $\mathcal A_{\mathrm{all}}(q)$ and the agreement $R(a_i,a^*_{all})$ with their majority-voting label $a^*_{\mathrm{all}}$ as a secondary supervision.

\paragraph{Progressive supervision.}
With training progress $s_t=t/T$, where $T$ is the total number of steps, the underlying supervision curriculum is
\begin{equation}
\begin{split}
r^{\mathrm{prog}}_{\mathrm{acc}}={}&(1-s_t)R(a_i,a^*_{\mathrm{all}})\\
&+s_tR(a_i,a^*_L).
\end{split}
\label{stepboost}
\end{equation}
Early updates rely more on all-response consensus, while later updates increasingly use LONG evidence. This schedule expresses a curriculum rather than a guarantee that LONG becomes correct as training proceeds.

\paragraph{Confidence weighting.}
Without ground-truth labels erroneous majority-vote pseudo-labels can steer policy updates toward incorrect answers \cite{ttrl}.
We therefore weight answer-agreement
reward $R(a_i,a^*_L)$ to reduce reliance on pseudo-labels, allowing varying degrees of supervision based on the ``confidence''.
Let $N_L=|\mathcal A_L|$, let $v_1,v_2$ be the largest and second-largest answer counts, and set $v_2=0$ when only one distinct answer occurs. For a minimum sample threshold $m\geq1$, define
\begin{equation}
c_L(q)=\begin{cases}(v_1-v_2)/N_L,&N_L\geq m,\\0,&N_L<m.\end{cases}
\label{eq:confidence}
\end{equation}
The final weight is $w_{q,t}=s_t c_L(q)$, giving
\begin{equation}
\begin{split}
r_{\mathrm{acc}}(q,a_i,t)={}&(1-w_{q,t})R(a_i,a^*_{\mathrm{all}})\\
&+w_{q,t}R(a_i,a^*_L).
\end{split}
\label{eq:accuracy_reward}
\end{equation}
Such a weighting makes insufficient \LONG{} samples or a tied vote return the answer reward to all-response consensus, avoiding rejecting the whole update. When the two pseudo-labels agree, the weight does not change this reward. We use this ``consensus strength'' as is a reliability ``confidence'', not a calibrated correctness probability, and a confidently wrong consensus can still reinforce errors.

\subsection{Behavioral Reward: SHORT Activation and Ratio Control}
\paragraph{From exploration to answer agreement.}
Answer-agreement rewards alone may favor the currently stronger LONG behavior, leaving SHORT with few training opportunities. We introduce a behavior bonus that encourages SHORT exploration early and increasingly emphasizes agreement with LONG consensus later.

Let $I_S(o_i)$ indicate a valid SHORT response. Its base behavior bonus is
\begin{equation}
b_i=I_S(o_i)\lambda_{\mathrm{br}}\big[(1-s_t)+s_tR(a_i,a^*_L)\big].
\label{eq:beh_base}
\end{equation}
A SHORT answer that disagrees with LONG consensus receives the decaying bonus $\lambda_{\mathrm{br}}(1-s_t)$, while an agreeing SHORT answer receives $\lambda_{\mathrm{br}}$, and other responses with wrong formats receive zero. This shifts learning from making SHORT available to improving the SHORT answers. Confidence weighting modifies Eq.~\eqref{eq:accuracy_reward}. The behavior bonus retains the time-based curriculum.
This time-based schedule the curriculum: first encourages \SHORT{} explorations and then leans to answer agreement.

\begin{figure}[t]
\centering
\includegraphics[width=\linewidth]{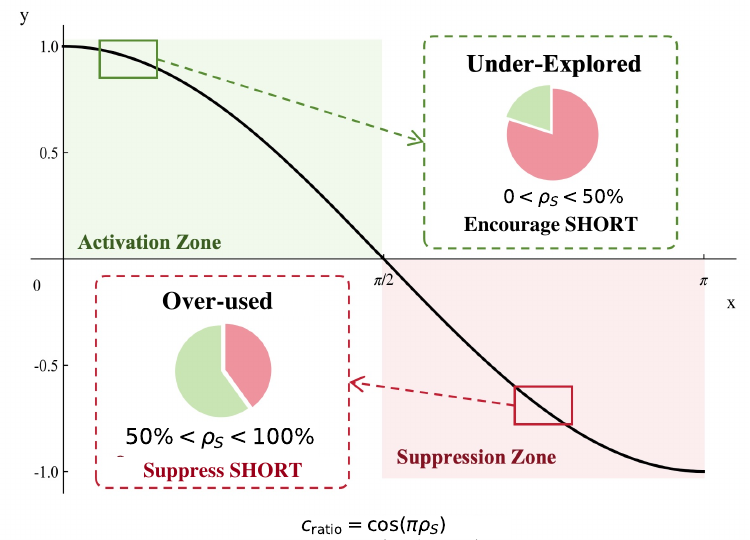}
\caption{Cosine ratio control for SHORT activation. The horizontal coordinate is $x=\pi\rho_S$ and the vertical coordinate is $c_{\mathrm{ratio}}$. Positive coefficients encourage SHORT exploration below $\rho_S=0.5$, while negative coefficients suppress further SHORT dominance above it. Pie charts illustrate mode proportions rather than measured distributions.}
\label{fig:ratio_controller}
\end{figure}

\paragraph{Regulating the mode mixture.}
Unrestricted activation can make SHORT dominate sampling, reducing LONG evidence and changing the all-response consensus. We compute the proportion $\rho_S$ of SHORT responses in the current training batch and rescale the bonus:
\begin{equation}
c_{\mathrm{ratio}}=\cos(\pi\rho_S),\qquad r_{\mathrm{beh}}=b_i c_{\mathrm{ratio}}.
\label{eq:ratio_coeff}
\end{equation}
When $\rho_S<0.5$, the bonus encourages SHORT. Above $0.5$, the negative coefficient opposes further SHORT dominance. Figure~\ref{fig:ratio_controller} illustrates this regulation: it does not constrain the learned mixture to remain at 50\% but change the direction of the bonus.

Together, the behavioral reward adjusts the ratio of \SHORT{} and \LONG{} reasoning mode, encouraging attempts at \SHORT{} while retaining \LONG{} to provide learning signals, thereby avoiding collapse into either extreme.
It provides batch-level feedback against mode imbalance, while answer quality still depends on consensus and policy optimization. 

\subsection{Joint Full-Trajectory Optimization}
\label{sec:optimization}
We use the combined reward in Eq.~\eqref{eq:total_reward} to optimize complete trajectories from both modes with GRPO. Writing $u_{i,k}(\theta)$ for the token probability ratio between the current and rollout policies, the clipped policy objective is
\begin{equation}
\begin{split}
J_{\mathrm{clip}}(\theta)=\mathbb E\Bigg[\frac1K\sum_{i=1}^{K}\frac1{|o_i|}\sum_{k=1}^{|o_i|}\min\Big(&u_{i,k}\widehat A_i,\\
\operatorname{clip}(u_{i,k},1-\delta,1+\delta)&\widehat A_i\Big)\Bigg].
\end{split}
\label{eq:grpo}
\end{equation}
Here $|o_i|$ counts the full generated trajectory for optimization. Both mode selection and subsequent generation receive updates. Algorithm~\ref{alg:pacereasoner} summarizes the learning procedure: no target-task reference answers, external teacher, or additional verifier is used. 

After adaptation, the policy generates a response conditioned on the question and the requested LONG or SHORT mode. SHORT generation uses the adapted policy directly, without requiring LONG responses or consensus computation.

\begin{algorithm}[t]
\caption{\our{} test-time adaptation}
\label{alg:pacereasoner}
\begin{algorithmic}[1]
\REQUIRE Initial policy $\pi_{\theta_0}$, questions $\mathcal D$, steps $T$
\FOR{$t=1,\ldots,T$}
\STATE Sample a question batch from $\mathcal D$
\STATE Generate $K$ mixed-mode responses per question
\STATE Parse modes and final answers
\STATE Compute all/LONG consensus and $c_L(q)$
\STATE Compute batch SHORT ratio $\rho_S$ and $s_t=t/T$
\STATE Compute answer reward using Eq.~\eqref{eq:accuracy_reward}
\STATE Compute behavior reward using Eqs.~\eqref{eq:beh_base}--\eqref{eq:ratio_coeff}
\STATE Add format reward and normalize within each group
\STATE Update both modes on their complete trajectories
\ENDFOR
\RETURN Adapted policy $\pi_\theta$
\end{algorithmic}
\end{algorithm}

\begin{table*}[t]
\centering
\small
\setlength{\tabcolsep}{5pt}
\begin{tabular}{ll*{10}{r}}
\toprule
\multirow{2}{*}{Mode} & \multirow{2}{*}{Method} & \multicolumn{2}{c}{CSQA} & \multicolumn{2}{c}{GPQA-D} & \multicolumn{2}{c}{MATH-500} & \multicolumn{2}{c}{GSM8K} & \multicolumn{2}{c}{Average} \\
\cmidrule(lr){3-4}\cmidrule(lr){5-6}\cmidrule(lr){7-8}\cmidrule(lr){9-10}\cmidrule(l){11-12}
& & Acc. & R-tok. & Acc. & R-tok. & Acc. & R-tok. & Acc. & R-tok. & Acc. & R-tok. \\
\midrule
\rowcolor{black!12}
\multicolumn{12}{l}{\textbf{Qwen3-4B-Base}} \\
\multirow{5}{*}{LONG} & Zero-shot & 50.9 & 413.4 & 21.3 & 461.5 & 50.0 & 510.0 & 60.4 & 328.8 & 45.7 & 428.4 \\
 & TTRL & \textbf{80.9} & 159.1 & 31.8 & 386.3 & \textbf{79.4} & \underline{501.2} & \underline{93.1} & 203.0 & \underline{71.3} & 312.4 \\
 & Length-Penalty & 39.9 & \underline{158.6} & \underline{33.3} & \underline{283.2} & 56.2 & 510.6 & 85.8 & 201.5 & 53.8 & \underline{288.5} \\
 & CRISP & 69.6 & 527.7 & 32.5 & 699.3 & 47.6 & 759.0 & 53.4 & \underline{186.6} & 50.8 & 543.1 \\
\rowcolor{black!5}
 & \our & \underline{80.8} & \textbf{92.0} & \textbf{36.9} & \textbf{168.0} & \underline{77.2} & \textbf{450.2} & \textbf{93.7} & \textbf{173.6} & \textbf{72.2} & \textbf{221.0} \\
\addlinespace[3pt]
\multirow{5}{*}{SHORT} & Zero-shot & 25.8 & 482.1 & 11.6 & 306.0 & \underline{31.6} & 401.0 & \underline{30.9} & 286.0 & 25.0 & 368.8 \\
 & TTRL & 78.1 & \underline{11.0} & 22.9 & 19.8 & 19.4 & \underline{13.4} & 22.9 & 19.8 & 35.8 & 16.0 \\
 & Length-Penalty & \textbf{80.7} & 12.4 & 14.0 & \underline{18.3} & 26.4 & 13.9 & 26.7 & \underline{15.6} & \underline{37.0} & \underline{15.1} \\
 & CRISP & 67.1 & \textbf{9.2} & \underline{31.1} & \textbf{9.2} & 23.4 & \textbf{7.8} & 21.2 & \textbf{13.2} & 35.7 & \textbf{9.8} \\
\rowcolor{black!5}
 & \our & \underline{80.2} & 12.0 & \textbf{36.9} & 51.4 & \textbf{39.8} & 14.5 & \textbf{39.1} & 19.3 & \textbf{49.0} & 24.3 \\
\midrule
\rowcolor{black!12}
\multicolumn{12}{l}{\textbf{Qwen3-8B-Base}} \\
\multirow{5}{*}{LONG} & Zero-shot & 77.0 & 149.8 & 29.8 & 514.0 & 61.8 & 466.0 & 72.1 & 228.3 & 60.2 & 339.5 \\
 & TTRL & \textbf{84.0} & \underline{127.4} & 31.3 & \underline{319.0} & \textbf{79.8} & 487.7 & \underline{94.2} & 210.5 & \underline{72.3} & \underline{286.2} \\
 & Length-Penalty & 61.4 & 133.7 & \underline{32.8} & 419.5 & 65.5 & \underline{463.6} & 86.8 & \textbf{205.3} & 61.6 & 305.5 \\
 & CRISP & 51.9 & 402.5 & 31.5 & 856.0 & 49.7 & 896.3 & 59.0 & 318.5 & 48.0 & 618.3 \\
\rowcolor{black!5}
 & \our & \underline{83.0} & \textbf{112.9} & \textbf{36.9} & \textbf{241.9} & \underline{77.8} & \textbf{462.9} & \textbf{95.0} & \underline{207.9} & \textbf{73.2} & \textbf{256.4} \\
\addlinespace[3pt]
\multirow{5}{*}{SHORT} & Zero-shot & 39.1 & 100.8 & 24.2 & 100.7 & \underline{34.6} & 206.0 & \underline{28.5} & 201.3 & 31.6 & 152.2 \\
 & TTRL & \underline{81.2} & 11.0 & \underline{31.3} & 11.0 & 27.2 & \underline{13.5} & 8.1 & 13.2 & 37.0 & \textbf{12.2} \\
 & Length-Penalty & 73.9 & 23.8 & 5.5 & 14.6 & 31.0 & \textbf{12.1} & 19.4 & \textbf{11.9} & 32.5 & 15.6 \\
 & CRISP & 68.4 & \underline{17.1} & 28.7 & \underline{11.2} & 26.8 & 13.9 & 25.0 & 14.8 & \underline{37.2} & 14.3 \\
\rowcolor{black!5}
 & \our & \textbf{82.5} & \textbf{11.0} & \textbf{36.4} & \textbf{11.0} & \textbf{37.2} & 14.0 & \textbf{46.0} & \underline{13.2} & \textbf{50.5} & \underline{12.3} \\
\midrule
\rowcolor{black!12}
\multicolumn{12}{l}{\textbf{Llama3.1-8B-Instruct}} \\
\multirow{5}{*}{LONG} & Zero-shot & 62.7 & 175.1 & 16.7 & 869.6 & 31.4 & 612.3 & 72.1 & 321.0 & 45.7 & 494.5 \\
 & TTRL & 66.9 & \underline{133.6} & \underline{27.9} & \underline{308.9} & 33.0 & 434.2 & \textbf{90.5} & 207.9 & \underline{54.6} & 271.2 \\
 & Length-Penalty & 60.4 & 166.8 & 23.2 & 354.5 & \textbf{35.9} & 417.1 & 84.4 & \underline{177.0} & 51.0 & 278.9 \\
 & CRISP & \underline{68.6} & \textbf{127.3} & 26.7 & 357.3 & 25.9 & \textbf{291.4} & 76.7 & \textbf{156.4} & 49.5 & \textbf{233.1} \\
\rowcolor{black!5}
 & \our & \textbf{68.7} & 145.2 & \textbf{28.8} & \textbf{214.6} & \underline{33.8} & \underline{400.2} & \underline{89.6} & 225.1 & \textbf{55.2} & \underline{246.3} \\
\addlinespace[3pt]
\multirow{5}{*}{SHORT} & Zero-shot & 60.6 & 187.1 & 14.4 & 790.8 & 10.6 & 634.7 & \underline{11.8} & 313.2 & 24.4 & 481.5 \\
 & TTRL & \underline{66.3} & \underline{11.0} & 20.1 & 30.7 & 1.0 & \underline{13.0} & 7.3 & 13.1 & 23.7 & \underline{17.0} \\
 & Length-Penalty & 63.5 & 16.3 & 23.5 & 37.6 & 6.4 & 69.7 & 8.5 & \underline{11.8} & 25.5 & 33.9 \\
 & CRISP & \textbf{68.6} & \textbf{4.0} & \textbf{29.5} & \textbf{4.9} & \underline{11.0} & \textbf{6.3} & 10.6 & \textbf{5.4} & \underline{29.9} & \textbf{5.2} \\
\rowcolor{black!5}
 & \our & 65.6 & 12.6 & \underline{28.8} & \underline{30.1} & \textbf{14.4} & 14.0 & \textbf{15.5} & 13.0 & \textbf{31.1} & 17.4 \\
\bottomrule
\end{tabular}
\caption{Performance comparison under LONG and SHORT reasoning modes for Qwen3-4B/8B-Base and Llama3.1-8B-Instruct models. Acc.: mean@8 accuracy (\%); R-tok.: reasoning tokens before the final answer, excluding answer tokens. Average is the unweighted mean across four tasks. Shading highlights \our. Bold and underlining indicate the best and second-best results within each backbone and mode, respectively.}
\label{tab:main_tab}
\end{table*}

\begin{table*}[t]
\centering\small
\begin{tabular}{lrrrrrrrr}
\toprule
&\multicolumn{2}{c}{CSQA}&\multicolumn{2}{c}{GPQA-D}&\multicolumn{2}{c}{MATH-500}&\multicolumn{2}{c}{GSM8K}\\
Supervision&Acc.&Time&Acc.&Time&Acc.&Time&Acc.&Time\\
\midrule
All-response&79.6&2h 24m&32.7&44m&30.2&2h 17m&26.1&4h 25m\\
LONG-only&78.9&2h 13m&28.2&44m&\textbf{45.0}&2h 22m&36.9&4h 38m\\
Progressive&\textbf{80.2}&1h 57m&\textbf{36.9}&40m&39.8&2h 9m&\textbf{39.1}&3h 51m\\
\bottomrule
\end{tabular}
\caption{Supervision ablation on Qwen3-4B-Base. Acc.: SHORT accuracy (\%); time: RL training time.}
\label{tab:gt_ablation}
\end{table*}

\begin{table*}[t]
\centering\footnotesize
\setlength{\tabcolsep}{1.25pt}
\begin{tabular}{@{}cccc@{\hspace{4pt}}rrr@{\hspace{4pt}}rrr@{\hspace{4pt}}rrr@{\hspace{4pt}}rrr@{}}
\toprule
\multirow{2}{*}{\shortstack{SHORT\\exploration}}&\multirow{2}{*}{\shortstack{LONG-answer\\matching}}&\multirow{2}{*}{\shortstack{SHORT ratio\\control}}&\multirow{2}{*}{\shortstack{Eval.\\mode}}&\multicolumn{3}{c}{CSQA}&\multicolumn{3}{c}{GPQA-D}&\multicolumn{3}{c}{MATH-500}&\multicolumn{3}{c}{GSM8K}\\
\cmidrule(lr){5-7}\cmidrule(lr){8-10}\cmidrule(lr){11-13}\cmidrule(lr){14-16}
&&&&Acc.&R-tok.&Score&Acc.&R-tok.&Score&Acc.&R-tok.&Score&Acc.&R-tok.&Score\\
\midrule
No&No&No&LONG&79.2&128.1&0.6&36.0&107.5&0.3&76.7&479.8&0.2&93.3&206.6&0.5\\
\midrule
Yes&No&Yes&SHORT&79.3&\underline{14.7}&5.4&\underline{33.3}&31.4&\textbf{1.1}&\textbf{40.0}&14.5&\textbf{2.8}&\underline{25.8}&19.3&1.3\\
Yes&Yes&No&SHORT&77.3&12.0&\underline{6.4}&0.0&\textbf{11.0}&0.0&21.4&14.5&1.5&16.8&19.3&0.9\\
Yes&No&No&SHORT&\underline{80.0}&12.0&6.7&22.7&\underline{31.4}&0.7&24.4&14.5&1.7&25.2&19.3&\underline{1.3}\\
Yes&Yes&Yes&SHORT&\textbf{80.2}&\textbf{12.0}&\textbf{6.7}&\textbf{36.9}&51.4&\underline{0.7}&\underline{39.8}&\textbf{14.5}&\underline{2.7}&\textbf{39.1}&\textbf{19.3}&\textbf{2.0}\\
\bottomrule
\end{tabular}
\caption{Behavior-reward ablation on Qwen3-4B-Base. SHORT exploration rewards choosing SHORT; LONG-answer matching rewards SHORT answers that match LONG consensus (Eq.~\eqref{eq:beh_base}). SHORT ratio control scales this bonus using the sampled SHORT proportion (Eq.~\eqref{eq:ratio_coeff}). The last row enables all three components. Score = Acc./R-tok.; accuracy is in percentage units. Among SHORT results, bold and underlining mark the best and second-best distinct reported values, respectively (higher Acc./Score, lower R-tok.). The LONG row is a reference.}
\label{tab:behavior_ablation}
\end{table*}

\begin{table*}[t]
\centering\small
\setlength{\tabcolsep}{6pt}
\renewcommand{\arraystretch}{1.15}
\begin{tabular}{llcccc}
\toprule
Reasoning mode&Model&CSQA&GPQA-D&MATH-500&GSM8K\\
\midrule
\multirow{4}{*}{LONG}&Qwen3-4B (Thinking)&56.8&15.2&65.4&92.3\\
&4B-Base (\our)&\textbf{80.8}&\textbf{36.9}&\textbf{77.2}&\textbf{93.7}\\
\cmidrule(lr){2-6}
&Qwen3-8B (Thinking)&82.0&16.7&61.4&93.6\\
&8B-Base (\our)&\textbf{83.0}&\textbf{36.9}&\textbf{77.8}&\textbf{95.0}\\
\midrule
\multirow{4}{*}{SHORT}&Qwen3-4B (Non-thinking)&76.7&\textbf{37.8}&18.2&18.4\\
&4B-Base (\our)&\textbf{80.2}&36.9&\textbf{39.8}&\textbf{39.1}\\
\cmidrule(lr){2-6}
&Qwen3-8B (Non-thinking)&79.4&\textbf{37.4}&24.2&21.1\\
&8B-Base (\our)&\textbf{82.5}&36.4&\textbf{37.2}&\textbf{46.0}\\
\bottomrule
\end{tabular}
\caption{Accuracy (\%) of \our{} adapted from Qwen3-4B/8B-Base and released Qwen3-4B/8B checkpoints. Thinking and non-thinking are paired with LONG and SHORT, respectively. Bold indicates the higher accuracy within each model-size and mode pair.}
\label{tab:appendix_baselines}
\end{table*}

\section{Experiments}
\label{experiments}
We evaluate whether \our{} improves SHORT-mode accuracy while reducing reasoning cost, and analyze the contribution of each method component.

\subsection{Experimental Setup}
\label{sec:setup}
\paragraph{Benchmarks.}
We evaluate on GSM8K and MATH-500 for mathematical reasoning
and verifiable answers \citep{gsm8k,math500}, and on CSQA and
GPQA-Diamond for commonsense and scientific reasoning
\citep{csqa,gpqa}. Together, these benchmarks cover both
open-ended and multiple-choice questions.

\paragraph{Implementation.}
We evaluate Qwen3-4B-Base, Qwen3-8B-Base, and Llama3.1-8B-Instruct \citep{qwen3technicalreport,llama}. Adaptation is performed independently on each target benchmark without reference-answer rewards. We use GRPO in EasyR1 \citep{shao2024deepseekmath,easyr1}, AdamW with learning rate $10^{-6}$, 15 epochs, and 15 rollouts per question at temperature 1.0. Rollout batches contain 256 questions for CSQA/GSM8K and 64 for GPQA-Diamond/MATH-500. The configured actor update batch sizes are respectively 64 and 16 question groups, corresponding to 960 and 240 response trajectories after expansion by 15 rollouts. These are distinct from rollout batch sizes. We use $\lambda_{\mathrm{br}}=2$ and 8 NVIDIA H20 GPUs.

\paragraph{Baselines.}
We compare zero-shot CoT, TTRL \citep{ttrl}, Length-Penalty, and CRISP \citep{crisp} from the same initial checkpoint for each backbone, using the same target data for adaptation and a unified mean@8 evaluation protocol. Length-Penalty is our adaptation of reasoning-length control ideas from ThinkPrune \citep{hou2025thinkprune}, DLER \citep{liu2025dler}, and L1 \citep{aggarwal2025l1} to the label-free target-task setting. Its reward is $R_{\mathrm{fmt}}+R_{\mathrm{acc}}R_{\mathrm{length}}$, where $R_{\mathrm{length}}=1/(L-L_{\mathrm{med}})$ for response length $L$ above the batch median $L_{\mathrm{med}}$, and 1 otherwise. 


\paragraph{Evaluation Metrics.}
For each question and mode, we sample eight responses and score
each final answer as correct (1) or incorrect (0). Accuracy is
the mean of these scores across responses and questions,
reported as mean@8 (\%). We do not use voting or best-of-eight
selection. R-tok. is the average number of generated reasoning
tokens before the final answer, excluding answer tokens. We
report unweighted averages across tasks. Table~\ref{tab:main_tab}
reports accuracy and reasoning cost separately for LONG and
SHORT. All main-table \our{} results use the final method with
confidence-weighted progressive supervision; the full-method
reference results in the component studies match the main table.

\subsection{Short Reasoning Performance}
\paragraph{Comparison with baselines.}
\our{} achieves higher average SHORT accuracy than all baselines, exceeding TTRL by 13.2, 13.5, and 7.4 points on Qwen3-4B, Qwen3-8B, and Llama3.1-8B at 24.3, 12.3, and 17.4 reasoning tokens, respectively (Table~\ref{tab:main_tab}). All three improve over zero-shot with fewer tokens. Qwen3-8B's cost is similar to TTRL (12.3 versus 12.2 tokens) and below CRISP and Length-Penalty. Llama also uses fewer tokens than Length-Penalty. Qwen3-8B gains 37.9 GSM8K points over TTRL. Both average 13.2 reasoning tokens. These comparisons use average costs, not matched per-question budgets.

LONG also achieves higher average accuracy than all baselines, using fewer tokens than TTRL and Length-Penalty on all backbones and than CRISP on both Qwen models. SHORT improvements thus accompany a stronger LONG mode.

\paragraph{LONG--SHORT accuracy gap.}
The absolute LONG--SHORT accuracy gap is within 1 point in 41.7\% (5/12) of model--task pairs and within 5 points in 50\% (6/12). The latter cover CSQA and GPQA-D across all three models, with SHORT using 69.4\%--95.5\% fewer tokens. This supports SHORT for timely, sufficiently accurate answers on these tasks, while larger mathematical gaps motivate retaining LONG.

\subsection{Ablation Study}
\label{sec:components}
Table \ref{tab:gt_ablation} and Table \ref{tab:behavior_ablation} present ablation results to analyze the contribution of method components.
\paragraph{Supervision strategy.}
Progressive supervision gives the highest SHORT accuracy on three tasks and the shortest training time on all four (Table~\ref{tab:gt_ablation}), supporting its accuracy and training-efficiency benefits. LONG-only nevertheless leads on MATH-500 (45.0\%), so supervision preferences remain task-dependent.

\paragraph{Behavior reward and ratio control.}
Table~\ref{tab:behavior_ablation} ablates SHORT exploration, LONG-answer matching, and ratio control. Score measures accuracy per reasoning token \citep{kaiser2026efficiency}. Without behavior rewards, the model relies on LONG. Removing ratio control reduces GSM8K accuracy from 39.1\% to 16.8\% at the same 19.3 tokens. Removing LONG-answer matching reduces GSM8K accuracy to 25.8\% and GPQA-D accuracy from 36.9\% to 33.3\%. GPQA-D length falls from 51.4 to 31.4 tokens, increasing Score from 0.7 to 1.1 despite lower accuracy. MATH-500 differs by only 0.2 points. The components thus provide task-dependent accuracy gains.

\subsection{Comparison with Post-trained Qwen3}
\label{sec:posttrained_reference}
To assess our lable-free test-time adaptation with post-training method, 
Table~\ref{tab:appendix_baselines} compares \our{} with the released Qwen3-4B/8B checkpoints, using \texttt{enable\_thinking=True/False} for their thinking/non-thinking modes \citep{qwen3technicalreport}. SHORT exceeds non-thinking accuracy on six of eight model--task pairs, gaining 13.0--24.9 points on mathematics. CSQA gains are smaller, at 3.1--3.5 points. GPQA-D is 0.9--1.0 points lower. LONG exceeds thinking accuracy on all eight pairs. This reference comparison involves models with different post-training histories.

\subsection{Cross-Dataset Transfer}
\label{sec:transfer}
\begin{table}[t]
\centering\small
\setlength{\tabcolsep}{5pt}
\begin{tabular}{llrr}
\toprule
Eval. dataset&Training data&LONG&SHORT\\
\midrule
\multirow{3}{*}{GPQA-D}&None (zero-shot)&21.3&11.6\\
&GPQA-D&36.9&36.9\\
&CSQA&37.4&34.7\\
\midrule
\multirow{3}{*}{MATH-500}&None (zero-shot)&50.0&31.6\\
&MATH-500&77.2&39.8\\
&GSM8K&74.2&36.8\\
\bottomrule
\end{tabular}
\caption{Cross-dataset accuracy (\%) on Qwen3-4B-Base. Each row specifies the evaluation and training datasets. None (zero-shot) uses the initial model; all trained rows use \our{} with unlabeled questions from the listed training dataset.}
\label{tab:generalization}
\end{table}

To investigate whether SHORT-mode improvements transfer beyond the
adaptation dataset,
we further adapt Qwen3-4B-Base on CSQA or GSM8K and evaluate on GPQA-D or MATH-500, respectively shown in Table~\ref{tab:generalization}. SHORT gains 23.1 and 5.2 points over zero-shot, trailing direct target-set adaptation by 2.2 and 3.0 points. This retains 63--91\% of the target-adaptation gains. LONG differs from target-set adaptation by +0.5 and $-3.0$ points. Both modes improve over zero-shot without target-set adaptation, suggesting that the gains are not solely due to same-dataset overfitting.

\section{Conclusion}

We presented \our{}, a label-free test-time reinforcement learning framework that improves SHORT-mode accuracy using answer supervision from the same model's LONG responses. Progressive, confidence-weighted supervision and behavior control coordinate answer learning and mode sampling within a shared policy. Across three models and four benchmarks, \our{} improves average SHORT accuracy over TTRL by 7.4--13.5 percentage points with fewer than 25 reasoning tokens per response on average. Ablations support the key components, and cross-dataset results demonstrate transfer beyond the adaptation set. These findings support using a model's own long-reasoning answers to strengthen its performance with short reasoning.

\FloatBarrier
\section*{Limitations}
Our evaluation covers three 4B/8B models and four question-answering benchmarks, without coding or interactive deployment tests. Target-set adaptation and two transfer experiments provide limited evidence of generalization. Multi-seed variability is not measured.

Self-generated consensus can reinforce shared errors and depends on the model retaining useful LONG reasoning capability. Confidence weighting is heuristic, and its incremental benefit is not isolated by the available ablations. Ratio control does not guarantee per-question LONG coverage; sensitivity to controller and reward settings remains a limitation. SHORT is a prompted mode rather than a fixed token budget.

Our cost metric counts reasoning tokens before the final answer, excluding answer tokens and adaptation computation. These measurements do not establish lower end-to-end latency, total deployment cost, or energy use.

\bibliography{Reference}
\clearpage
\appendix
\section{Additional Experimental Details}
\label{appendix}
\subsection{Benchmarks and Data}
\label{appendix_bench}
CSQA evaluates commonsense question answering with five answer options. GPQA-Diamond contains 198 graduate-level science questions. MATH-500 contains 500 problems from MATH, while GSM8K evaluates arithmetic word problems \citep{csqa,gpqa,math500,gsm8k}. The supplied question pools contain 1,221 CSQA, 198 GPQA-Diamond, 500 MATH-500, and 1,319 GSM8K items. For each task, adaptation and evaluation files contain the same question sequence. We use these questions as unlabeled target data for independent adaptation and evaluate their final answers. MATH-500 does not require a particular visible reasoning length for an answer to be scored correct. Multiple-choice and mathematical answers are parsed and normalized for voting and evaluation.

\paragraph{Evaluation generation settings.}
\label{appendix_generation}
Zero-shot, TTRL, and \our{} use forced LONG and SHORT instructions. For CRISP, the LONG group reports its native short-reasoning setting; the SHORT group reports native no-think for Qwen3-8B-Base and answer-only generation for the other backbones. These results use the unified LONG/SHORT labels in Table~\ref{tab:main_tab}; actual reasoning lengths are measured for each method.

\subsection{Training Configuration}
\label{appendix_setup}
Table~\ref{tab:training_parameters_full} distinguishes a rollout batch of questions from an actor update batch. With 15 responses per question, the training framework expands the configured actor batch by a factor of 15 before distributing it across workers. Thus, a configured actor batch of 64 denotes 960 response trajectories globally, not 64 trajectories per GPU. Main-table evaluation uses the unified mean@8 protocol described in Section~\ref{sec:setup}.

\begin{table*}[t]
\centering\small
\begin{tabular}{lcc}
\toprule
Parameter & CSQA / GSM8K & GPQA-D / MATH-500\\
\midrule
Rollout batch (questions) &256&64\\
Actor update batch (question groups, configured)&64&16\\
Responses per question&15&15\\
Rollout batch (response trajectories)&3840&960\\
Actor update batch (response trajectories)&960&240\\
Training epochs&15&15\\
Learning rate&$10^{-6}$&$10^{-6}$\\
Optimizer&AdamW&AdamW\\
Weight decay&0.01&0.01\\
Learning-rate warmup ratio&0&0\\
Rollout temperature&1.0&1.0\\
Maximum prompt / response length&2048 / 2048&2048 / 2048\\
Reference KL loss coefficient&0.01&0.01\\
Behavior bonus $\lambda_{\mathrm{br}}$&2&2\\
GPUs / tensor parallel size&8 / 2&8 / 2\\
Actor update microbatch (responses per device)&1&1\\
Gradient checkpointing&Enabled&Enabled\\
Rollout GPU memory utilization&0.6&0.6\\
\bottomrule
\end{tabular}
\caption{Training settings. Batch quantities are global unless explicitly marked per device. Maximum response length is an overall generation limit, not a separate SHORT budget.}
\label{tab:training_parameters_full}
\end{table*}

\subsection{Consensus Construction}
\label{appendix_majority_voting}
For each question, final-answer extraction forms an all-response answer pool and a LONG-only pool. We vote over normalized final answers, rather than entire reasoning traces. If a pool is empty, its pseudo-label is missing and its agreement reward is zero. A tie in the two leading LONG answer counts yields zero confidence in Eq.~\eqref{eq:confidence}; this affects the answer-supervision weight but does not remove the time-based behavior bonus. The additional confidence weight therefore reduces reliance on weak LONG consensus without establishing correctness of the fallback consensus.

Algorithm~\ref{alg:majority_voting} retains the original construction of the LONG-only (\texttt{think}) and all-response answer pools. Confidence weighting is applied after this voting step.
\begin{algorithm}[tb]
  \caption{Majority Voting for Pseudo Ground Truth (Reasoning / All Rollouts)}
  \label{alg:majority_voting}
  \begin{algorithmic}
    \STATE {\bfseries Input:} rollout set $\mathcal{O}(q)=\{o_i\}_{i=1}^{N}$ for each query $q$
    \STATE {\bfseries Output:} pseudo ground truths $gt_{\text{think}}(q)$ and $gt_{\text{all}}(q)$
    \STATE Initialize dictionaries $\mathcal{A}_{\text{think}} \leftarrow \{\}$, $\mathcal{A}_{\text{all}} \leftarrow \{\}$
    \FOR{each response $o_i \in \mathcal{O}(q)$}
        \STATE $o_i \leftarrow \textsc{StripSpacesAroundTags}(o_i)$
        \STATE $m_i \leftarrow \textsc{DetectMode}(o_i)$
        \STATE $a_i \leftarrow \textsc{ExtractAnswer}(o_i)$
        \IF{$a_i \neq \varnothing$}
            \STATE Append $a_i$ to $\mathcal{A}_{\text{all}}[q]$
            \IF{$m_i = \texttt{think}$}
                \STATE Append $a_i$ to $\mathcal{A}_{\text{think}}[q]$
            \ENDIF
        \ENDIF
    \ENDFOR
    \FOR{each query $q$}
        \IF{$|\mathcal{A}_{\text{think}}[q]| > 0$}
            \STATE $gt_{\text{think}}(q) \leftarrow \arg\max\limits_{a \in \mathcal{A}_{\text{think}}[q]} \; \textsc{Count}(a,\mathcal{A}_{\text{think}}[q])$
        \ELSE
            \STATE $gt_{\text{think}}(q) \leftarrow \varnothing$
        \ENDIF
        \IF{$|\mathcal{A}_{\text{all}}[q]| > 0$}
            \STATE $gt_{\text{all}}(q) \leftarrow \arg\max\limits_{a \in \mathcal{A}_{\text{all}}[q]} \; \textsc{Count}(a,\mathcal{A}_{\text{all}}[q])$
        \ELSE
            \STATE $gt_{\text{all}}(q) \leftarrow \varnothing$
        \ENDIF
    \ENDFOR
  \end{algorithmic}
\end{algorithm}

\subsection{Mode Instructions}
\label{Position}
The following template states the LONG/SHORT generation contract used in the revised method description. At forced-mode evaluation, the corresponding opening marker conditions generation. A direct answer is allowed as the empty-reasoning special case of SHORT; there is no separate SHORT token cap.
\begin{promptbox}{LONG/SHORT mode specification}
Choose a solution mode for the question.
LONG: develop the necessary explicit analysis,
intermediate derivations, and checks. Use:
<reason>reasoning</reason><answer>final answer</answer>
SHORT: solve concisely and directly, retaining only
essential steps when needed. Use:
<not>brief reasoning</not><answer>final answer</answer>
A direct answer may use an empty <not></not> block.
Place only the final answer in <answer>...</answer>.
\end{promptbox}

\subsection{Recorded Training Prompt}
\label{app:training-prompt}
The recorded training prompt below uses the \texttt{<not></not>} marker for the direct-answer case of SHORT.
\begin{promptbox}{Recorded training prompt}
You are required to first decide whether to think or not based on the given question. The reasoning process MUST BE enclosed within <reason> </reason> tags, and the final answer MUST BE enclosed within <answer> </answer> tags.
# Output Format
- If you decide NOT to think, you should ONLY output <not></not> and DIRECTLY output the final answer, and DO NOT contain any thinking process, the final answer MUST BE enclosed within <answer> </answer> tags. Output as '<not></not><answer> </answer>'. ONLY include the final answer between '<answer> </answer>' tags and do not include any other texts.
- If you decide to think, then you MUST enclose thinking process within <reason> </reason> tags, and the final answer MUST BE enclosed within <answer> </answer> tags. Output as '<reason> </reason><answer> </answer>'. Do not contain any <answer> </answer> tags within <reason> </reason> tokens, and all the texts MUST BE included in tags '<reason> </reason>' and '<answer> </answer>'.
# Requirements
ALL the output texts MUST BE enclosed within tags.
\end{promptbox}

\section{Additional Component Results}
\label{more_ablations}
The tables in this section report Qwen3-4B-Base component studies, with full-method reference results matching Table~\ref{tab:main_tab}. Gray entries identify runs evaluated using LONG after failing to activate the intended short behavior, and should not be read as SHORT measurements.

\subsection{Controller Shape and Bonus Strength}
Table~\ref{tab:ratio_controller_ablation} compares the controller functions, and Table~\ref{tab:ratio_controller_shifts_ablation} retains the reported shifted-cosine settings. These choices differ in their sign, scale, and behavior near the endpoints, so the comparison does not isolate smoothness or establish a unique optimal controller. The unshifted cosine gives the highest SHORT accuracy among the controller settings on all four tasks, although its higher GPQA accuracy comes with longer reasoning and a lower accuracy-per-token score. In the bonus-strength comparison (Table~\ref{tab:ratio_controller_ablation_behavior_bonus}), a value of 1 fails to activate SHORT on the two mathematical tasks; a value of 2 gives the highest SHORT accuracy on three tasks, while 3 is slightly better on GSM8K (39.3\% versus 39.1\%) with more reasoning tokens. On GPQA, the default bonus of 2 improves accuracy at the cost of longer reasoning, rather than a higher accuracy-per-token score.

\begin{table*}[htbp]
  \centering
  \small

  \setlength{\tabcolsep}{4.5pt}
  \renewcommand{\arraystretch}{1.15}

  \resizebox{\textwidth}{!}{%
  \begin{tabular}{l ccc|ccc|ccc|ccc}
    \toprule
    \textbf{Ratio Controller} &
    \multicolumn{3}{c|}{\textbf{CSQA}} &
    \multicolumn{3}{c|}{\textbf{GPQA}} &
    \multicolumn{3}{c|}{\textbf{MATH-500}} &
    \multicolumn{3}{c}{\textbf{GSM8K}} \\
    \cmidrule(lr){2-4}\cmidrule(lr){5-7}\cmidrule(lr){8-10}\cmidrule(lr){11-13}
    $f(x)$ & acc & R-tok. & score
    & acc & R-tok. & score
    & acc & R-tok. & score
    & acc & R-tok. & score \\
    \midrule

    $1$
      & 77.3 & 12.0 & 6.4
      & 0.0  & 11.0 & 0.0
      & 21.4 & 14.5 & 1.5
      & 16.8 & 19.3 & 0.9 \\

    $1-x$
      & 78.5 & 11.6 & 6.8
      & 20.7 & 31.5 & 0.7
      & 32.5 & 14.4 & 2.3
      & 36.0 & 19.3 & 1.9 \\

    $\frac{1}{2} - x$
      & 79.8 & 12.0 & 6.7
      & 31.2 & 31.9 & 1.0
      & 27.8 & 14.5 & 1.9
      & 26.0 & 19.1 & 1.4 \\

     $\frac{1}{x}$
      & 76.7 & 12.0 & 6.4
      & 22.2 & 31.5 & 0.7
      & 31.2 & 14.4 & 2.2
      & 37.2 & 19.3 & 1.9 \\

    $\cos(\pi x)$
      & 80.2 & 12.0 & 6.7
      & 36.9 & 51.4 & 0.7
      & 39.8 & 14.5 & 2.7
      & 39.1 & 19.3 & 2.0 \\

    \bottomrule
  \end{tabular}%
  }
\caption{Controller-function sensitivity on Qwen3-4B-Base. The reported score is accuracy divided by reasoning tokens; accuracy and length remain the primary measurements.}
\label{tab:ratio_controller_ablation}
\end{table*}

\begin{table*}[htbp]
  \centering
  \small

  \setlength{\tabcolsep}{4.5pt}
  \renewcommand{\arraystretch}{1.15}

  \resizebox{\textwidth}{!}{%
  \begin{tabular}{l ccc|ccc|ccc|ccc}
    \toprule
    \textbf{Ratio Controller} &
    \multicolumn{3}{c|}{\textbf{CSQA}} &
    \multicolumn{3}{c|}{\textbf{GPQA}} &
    \multicolumn{3}{c|}{\textbf{MATH-500}} &
    \multicolumn{3}{c}{\textbf{GSM8K}} \\
    \cmidrule(lr){2-4}\cmidrule(lr){5-7}\cmidrule(lr){8-10}\cmidrule(lr){11-13}
    $f(x)$ & acc & R-tok. & score
    & acc & R-tok. & score
    & acc & R-tok. & score
    & acc & R-tok. & score \\
    \midrule

    $\cos\!\big(\pi(x-\frac{\pi}{4})\big)$
      & 74.1 & 11.5 & 6.4
      & 28.8 & 31.6 & 0.9
      & 25.2 & 13.8 & 1.8
      & \g{90.3} & \g{198.3} & \g{0.5} \\

    $\cos\!\big(\pi(x+\frac{\pi}{4})\big)$
      & 77.2 & 12.0 & 6.4
      & 27.8 & 31.4 & 0.9
      & 32.3 & 13.5 & 2.4
      & \g{89.3} & \g{202.4} & \g{0.4} \\

    $\cos(\pi x)$
      & 80.2 & 12.0 & 6.7
      & 36.9 & 51.4 & 0.7
      & 39.8 & 14.5 & 2.7
      & 39.1 & 19.3 & 2.0 \\

    \bottomrule
  \end{tabular}%
  }
\caption{Shifted-controller sensitivity. Functions and values are retained as reported. Gray entries are LONG evaluations, not successful SHORT results.}
\label{tab:ratio_controller_shifts_ablation}
\end{table*}

\begin{table*}[htbp]
  \centering
  \small

  \setlength{\tabcolsep}{4.5pt}
  \renewcommand{\arraystretch}{1.15}

  \resizebox{\textwidth}{!}{%
  \begin{tabular}{l ccc|ccc|ccc|ccc}
    \toprule
    \textbf{Behavior Bonus} &
    \multicolumn{3}{c|}{\textbf{CSQA}} &
    \multicolumn{3}{c|}{\textbf{GPQA}} &
    \multicolumn{3}{c|}{\textbf{MATH-500}} &
    \multicolumn{3}{c}{\textbf{GSM8K}} \\
    \cmidrule(lr){2-4}\cmidrule(lr){5-7}\cmidrule(lr){8-10}\cmidrule(lr){11-13}
    $\lambda_{br}$ & acc & R-tok. & score
    & acc & R-tok. & score
    & acc & R-tok. & score
    & acc & R-tok. & score \\
    \midrule

    $1$
      & 79.9 & 13.0 & 6.1
      & 32.3 & 31.3 & 1.0
      & \g{78.0} & \g{440.7} & \g{0.2}
      & \g{92.2} & \g{194.8} & \g{0.5} \\

    $2$
      & 80.2 & 12.0 & 6.7
      & 36.9 & 51.4 & 0.7
      & 39.8 & 14.5 & 2.7
      & 39.1 & 19.3 & 2.0 \\

    $3$
      & 80.0 & 13.0 & 6.2
      & 30.3 & 35.0 & 0.9
      & 39.6 & 14.8 & 2.7
      & 39.3 & 21.9 & 1.8 \\

    \bottomrule
  \end{tabular}%
  }
\caption{Behavior-bonus sensitivity. Gray entries are LONG evaluations. The score is accuracy divided by reasoning tokens.}
\label{tab:ratio_controller_ablation_behavior_bonus}
\end{table*}

\subsection{Training Dynamics}
\label{sec:training_dynamics}
\begin{figure*}[t]
\centering
\begin{subfigure}[b]{0.29\textwidth}
\includegraphics[width=\linewidth]{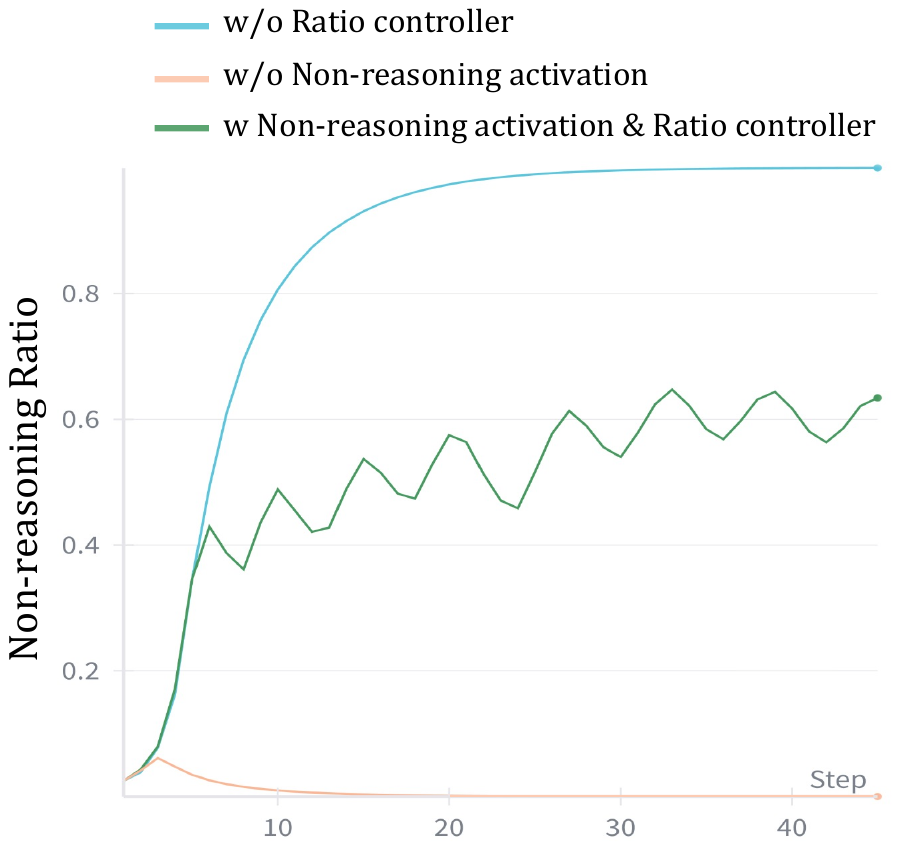}
\caption{SHORT ratio on GPQA.}
\end{subfigure}\hfill
\begin{subfigure}[b]{0.29\textwidth}
\includegraphics[width=\linewidth]{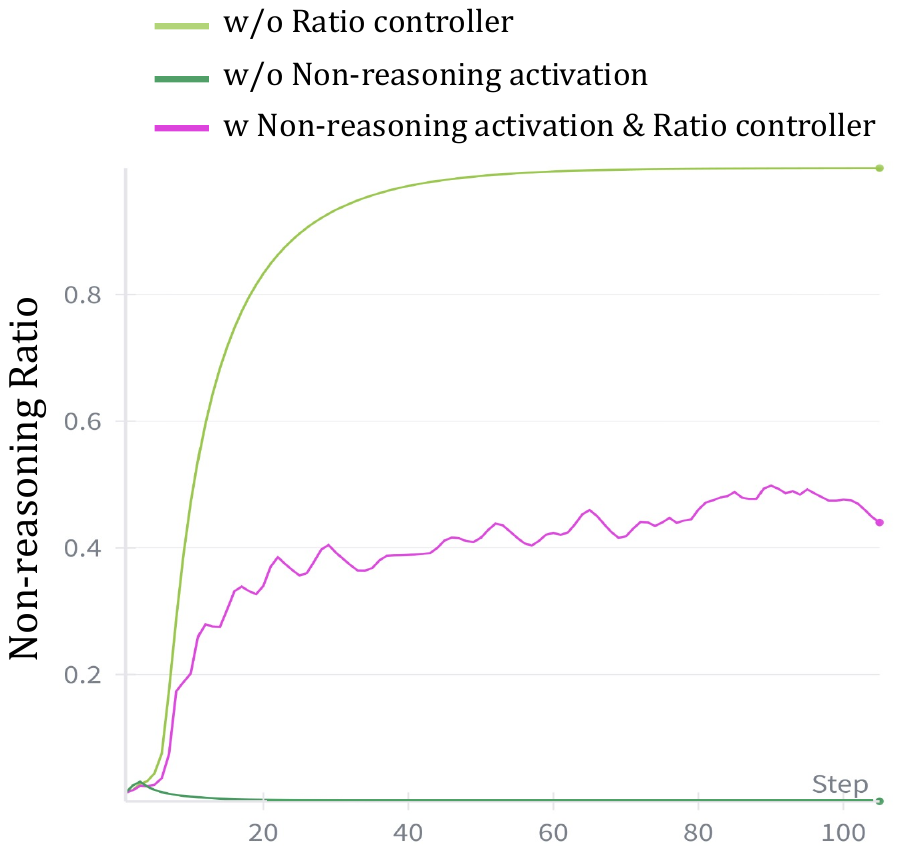}
\caption{SHORT ratio on MATH-500.}
\end{subfigure}\hfill
\begin{subfigure}[b]{0.39\textwidth}
\includegraphics[width=\linewidth]{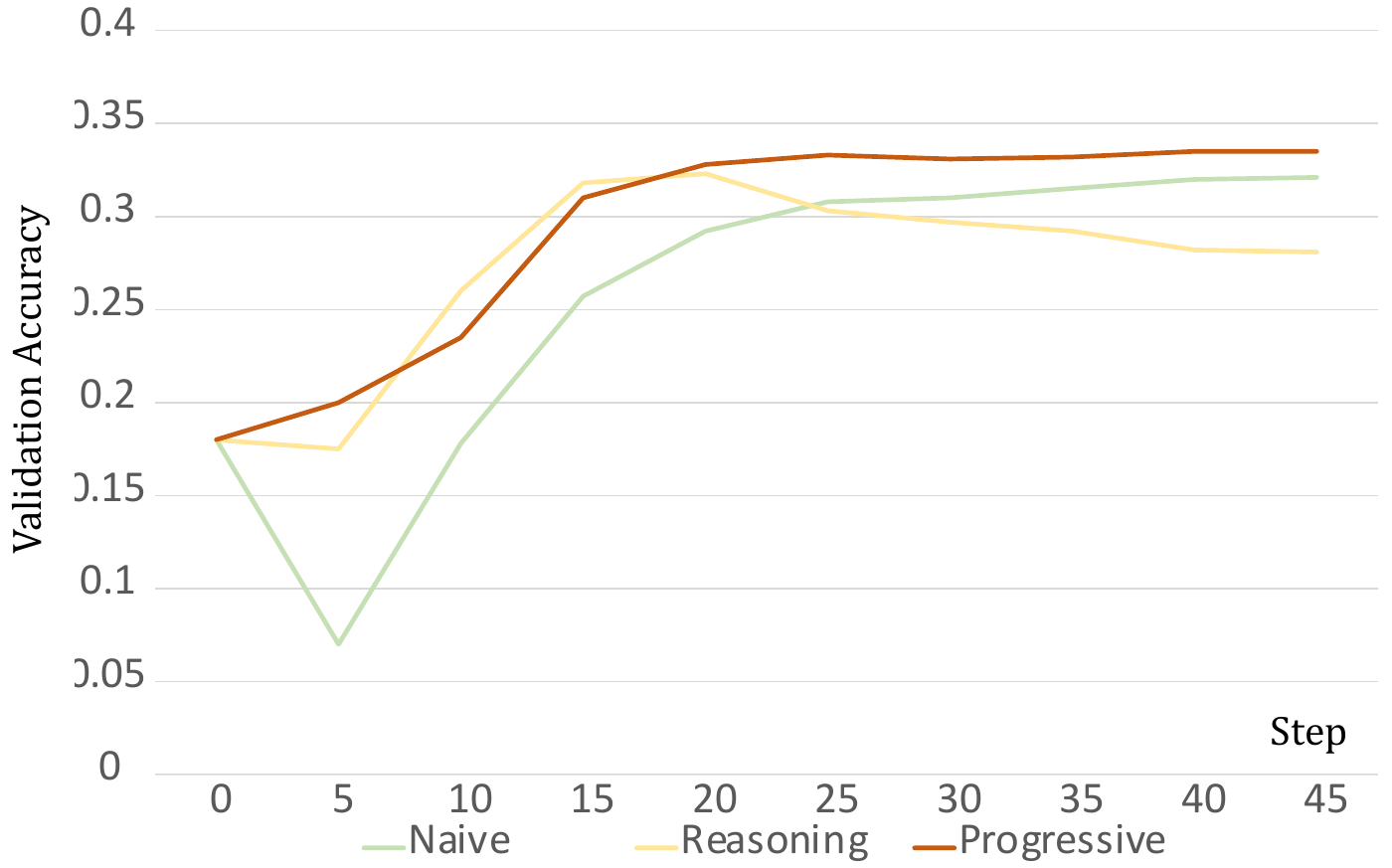}
\caption{Supervision on GPQA.}
\label{fig:supervision_dynamics}
\end{subfigure}
\caption{Training dynamics on Qwen3-4B-Base. (a--b) SHORT sampling with and without activation or ratio control; ``Non-reasoning'' denotes the direct-answer SHORT setting. (c) Validation accuracy under all-response (Naive), LONG-only (Reasoning), and progressive supervision.}
\label{fig:futher_ablation_2}
\end{figure*}

Figure~\ref{fig:futher_ablation_2}(a--b) shows that SHORT sampling approaches zero without activation and approaches one without ratio control. With both components, the rollout pool retains a mixture of modes, supplying SHORT training examples alongside LONG answers for supervision. These trajectories complement the behavior-reward ablation in Table~\ref{tab:behavior_ablation}.

On GPQA, all-response voting suffers an early accuracy drop and LONG-only voting declines later, whereas progressive supervision improves more steadily (Figure~\ref{fig:supervision_dynamics}). These trajectories complement the supervision comparison in Table~\ref{tab:gt_ablation}; they do not separately isolate the additional effect of confidence weighting.

\section{Recorded Qualitative Examples}
\label{more_qualitative_examples}
We retain the original recorded outputs for CSQA, GPQA, MATH-500, and GSM8K below. Their generated text is unchanged; these selected examples illustrate behavior and are not additional measurements from the updated main-table configurations. Mode markers inside a recorded response are preserved as part of that response. GT denotes the ground-truth reference answer. Table~\ref{tab:qual_examples_text} includes the original questions and SHORT-mode responses, including zero-shot outputs. The remaining tables retain the LONG-mode examples.

\begin{table*}[t]
\centering
\caption{Recorded SHORT-mode responses of Qwen3-4B-Base. The questions, reference answers, and generated outputs are preserved from the original examples.}
\small
\setlength{\tabcolsep}{6pt}
\renewcommand{\arraystretch}{1.2}
\begin{tabular}
{@{}p{0.09\textwidth}p{0.20\textwidth}p{0.20\textwidth}p{0.20\textwidth}p{0.20\textwidth}@{}}
\toprule
 & \textbf{CSQA} & \textbf{GPQA} & \textbf{MATH-500} & \textbf{GSM8K} \\
\midrule

\textbf{Question} &
\parbox[t]{\linewidth}{
\textbf{Question:} The man closed his eyes as the music played, what effect did the music have? (A)coma  (B)enjoyable  (C)soothing  (D)universal  (E)good or bad
} &
\parbox[t]{\linewidth}{
{\textbf{Question:} Astronomers are observing a planet with a radius of 1 Earth radius in a circular orbit, ... What is the maximum orbital period of a second planet, with a radius of 2.5 times that of Earth, ... A.37.5 B.33.5 C.12.5 D.7.5}
} &
\parbox[t]{\linewidth}{
\textbf{Question:} Compute $\frac{1}{\cos^2 10^\circ}+\frac{1}{\sin^2 20^\circ}+\frac{1}{\sin^2 40^\circ}$ 
} &
\parbox[t]{\linewidth}{
\textbf{Question:}Two trains leave San Rafael at the same time. They begin traveling westward, both traveling for 80 miles. The next day, they travel northwards, covering 150 miles. What's the distance covered by each train in the two days? 
}
\\
\midrule

\textbf{GT} &
\parbox[t]{\linewidth}{C} &
\parbox[t]{\linewidth}{B} &
\parbox[t]{\linewidth}{12} &
\parbox[t]{\linewidth}{230}
\\
\midrule

\textbf{Zero-shot's outputs} &
\parbox[t]{\linewidth}{The man closed his eyes as the music played, what effect did the music have? \texttt{<not>}\allowbreak\texttt{</not>}\allowbreak\texttt{<answer>}\allowbreak(C)soothing\allowbreak\texttt{</answer>}} & 
\parbox[t]{\linewidth}{To solve this problem, we need to determine the maximum orbital period of the second planet that can both transit and occult the first planet. Here's the step-by-step reasoning: ... \allowbreak\texttt{<answer>}\allowbreak D\allowbreak\texttt{</answer>}\allowbreak} & 
\parbox[t]{\linewidth}{To compute the expression $\frac{1}{\cos^2 10^\circ}+\frac{1}{\sin^2 20^\circ}+\frac{1}{\sin^2 40^\circ}$, we can use trigonometric identities and properties. First, recall the identity: $\frac{1}{\cos^2 \theta} = \sec^2 \theta,
\allowbreak\quad
\frac{1}{\sin^2 \theta} = \csc^2 \theta.$ 
... (no answer)
} &
\parbox[t]{\linewidth}{The total distance covered by each train in the two days is calculated by adding the distance traveled on the first day (80 miles) and the distance traveled on the second day (150 miles). This gives us a total distance of 230 miles.}
\\
\midrule

\textbf{TTRL's outputs} &
\parbox[t]{\linewidth}{\texttt{<not>}\allowbreak\texttt{</not>}\allowbreak\texttt{<answer>}\allowbreak C\texttt{</answer>}
} &
\parbox[t]{\linewidth}{\texttt{<not>}\allowbreak\texttt{</not>}\allowbreak\texttt{<answer>}\allowbreak C\texttt{</answer>}
} &
\parbox[t]{\linewidth}{\texttt{<not>}\allowbreak\texttt{</not>}\allowbreak\texttt{<answer>}\allowbreak 16\texttt{</answer>}} &
\parbox[t]{\linewidth}{\texttt{<not>}\allowbreak\texttt{</not>}\allowbreak\texttt{<answer>}\allowbreak 460 miles\texttt{</answer>}
}
\\
\midrule

\textbf{Ours} &
\parbox[t]{\linewidth}{\texttt{<not>}\allowbreak\texttt{</not>}\allowbreak\texttt{<answer>}\allowbreak C\texttt{</answer>}} &
\parbox[t]{\linewidth}{\texttt{<not>}\allowbreak\texttt{</not>}\allowbreak\texttt{<answer>}\allowbreak B.33.5\texttt{</answer>}} &
\parbox[t]{\linewidth}{\texttt{<not>}\allowbreak\texttt{</not>}\allowbreak\texttt{<answer>}\allowbreak 12\texttt{</answer>}} &
\parbox[t]{\linewidth}{\texttt{<not>}\allowbreak\texttt{</not>}\allowbreak\texttt{<answer>}\allowbreak 230\texttt{</answer>}
}
\\
\bottomrule
\end{tabular}
\label{tab:qual_examples_text}
\end{table*}

\begin{table*}[t]
\centering

\footnotesize
\setlength{\tabcolsep}{6pt}
\renewcommand{\arraystretch}{1.2}

\begin{tabular}
{@{}p{0.14\textwidth}p{0.82\textwidth}@{}}
\toprule
& \textbf{CSQA}\\
\midrule

\textbf{Question} &
\parbox[t]{\linewidth}{
{\textbf{Question:} A revolving door is convenient for two direction travel, but it also serves as a security measure at a what? (A)bank (B)library (C)department store (D)mall (E)new york}
}
\\
\midrule

\textbf{GT} &
\parbox[t]{\linewidth}{A}
\\
\midrule

\textbf{Zero-shot's outputs} &
\parbox[t]{\linewidth}{To determine whether to think or not, I will analyze the question and the options provided. **Question:** A revolving door is convenient for two-direction travel, but it also serves as a security measure at a what? **Options:** (A) bank (B) library (C) department store (D) mall (E) New York **Reasoning:** A revolving door is commonly used in places where a high volume of people move in and out, and it helps to prevent the spread of germs and maintain a sense of security. This is particularly important in financial institutions like banks, where security is a top priority. Given this information, I can conclude that the correct answer is: \texttt{<answer>}\allowbreak (A) bank\allowbreak\texttt{</\allowbreak answer>}}
\\
\midrule

\textbf{TTRL's outputs} &
\parbox[t]{\linewidth}{\texttt{<reason>}\allowbreak The question asks which location a revolving door serves as a security measure. A revolving door is commonly used as a security measure at banks to control access and prevent unauthorized entry. Libraries, department stores, and malls may use revolving doors for convenience, but banks are the most likely to use them for security purposes.\texttt{</reason>}\allowbreak\texttt{<answer>}\allowbreak A\allowbreak\texttt{</\allowbreak answer>}
} 
\\
\midrule

\textbf{\our{} 's outputs} &
\parbox[t]{\linewidth}{\texttt{<reason>}\allowbreak A revolving door is typically used as a security measure at banks, as it helps prevent tailgating (where one person follows another into a secure area) and ensures that only one person can pass through at a time. While revolving doors can also be found in other places like department stores and malls, the primary security function is most commonly associated with banks.\texttt{</reason>}\allowbreak\texttt{<answer>}\allowbreak A\allowbreak\texttt{</\allowbreak answer>}}
\\
\bottomrule
\end{tabular}
\caption{\textbf{Illustration of generated responses under LONG mode} on CSQA, including zero-shot, TTRL and \our. The base model is Qwen3-4B-Base.}
\label{tab:complete_qual_examples_reasoning_csqa}
\end{table*}

\begin{table*}[t]
\centering

\footnotesize
\setlength{\tabcolsep}{6pt}
\renewcommand{\arraystretch}{1.2}

\begin{tabular}
{@{}p{0.14\textwidth}p{0.82\textwidth}@{}}
\toprule
& \textbf{GPQA}\\
\midrule

\textbf{Question} &
\parbox[t]{\linewidth}{
{\textbf{Question:} You are analyzing a small peptidic compound that has been chemically synthesized. The 1H NMR spectrum of the crude compound appears consistent with the expected molecule, with the exception of two peaks that both correspond to the same alpha-proton. These two peaks have similar chemical shifts and roughly equal integrals (together they integrate to 1H - the expected integral); based on the coupling pattern, you can rule out spin-spin coupling as an explanation for the duplicate peaks. LC-MS analysis of the crude compound at elevated temperature shows two clearly defined peaks of equal intensities. Both peaks have the same mass spectrum, which is consistent with the expected molecule. What is the most likely explanation for these observations? A. The compound is contaminated with a precursor B. The crude compound exists as a mixture of diastereoisomers C. 'Double coupling' has occurred during an amide-bond forming reaction D. The crude compound exists as a mixture of enantiomers. You are required to first decide whether to think or not based on the given question.}
}
\\
\midrule

\textbf{GT} &
\parbox[t]{\linewidth}{B}
\\
\midrule

\textbf{Zero-shot's outputs} &
\parbox[t]{\linewidth}{You are a helpful assistant. \texttt{<reason>}\allowbreak The question describes a small peptidic compound with a 1H NMR spectrum showing two peaks for the same alpha-proton, which are consistent in chemical shift and integral. LC-MS analysis at elevated temperature shows two peaks of equal intensities with the same mass spectrum. The absence of spin-spin coupling and the consistency with the expected molecule suggest that the issue is not due to contamination or diastereoisomers. The most plausible explanation is that the compound exists as a mixture of enantiomers, as enantiomers have identical NMR spectra and mass spectra but different optical rotations. The observation of two peaks in the NMR spectrum could be due to the presence of both enantiomers in the sample.\texttt{</reason>}\allowbreak \allowbreak\texttt{<answer>}\allowbreak D\allowbreak\texttt{</\allowbreak answer>}}
\\
\midrule

\textbf{TTRL's outputs} &
\parbox[t]{\linewidth}{\texttt{<reason>}\allowbreak The question describes a scenario where a synthetic peptidic compound shows two peaks in its 1H NMR spectrum that correspond to the same alpha-proton, with similar chemical shifts and integrals. The LC-MS analysis at elevated temperature shows two peaks of equal intensity with the same mass spectrum. The key points are: 1. **NMR Analysis**: Two peaks for the same proton, ruling out spin-spin coupling. 2. **LC-MS Analysis**: Two peaks of equal intensity with the same mass spectrum, indicating identical molecular species. The most likely explanation for these observations is that the compound exists in a form where these two peaks arise from different molecular conformations or isomers. Since the peaks have similar chemical shifts and integrals, they are likely not due to contamination or precursor molecules. Double coupling is not a plausible explanation given the context. Enantiomers would typically show distinct peaks in NMR due to their chirality, but here the peaks are identical. Diastereoisomers, however, can have similar NMR spectra due to their structural similarities, especially if they are in a conformational equilibrium.\texttt{</reason>}\allowbreak\texttt{<answer>}\allowbreak B. The crude compound exists as a mixture of diastereoisomers \allowbreak\texttt{</\allowbreak answer>}
} 
\\
\midrule

\textbf{\our{} 's outputs} &
\parbox[t]{\linewidth}{\texttt{<reason>}\allowbreak The question describes a situation where a synthetic peptidic compound shows two peaks in its 1H NMR spectrum that appear to be the same alpha-proton but do not couple to each other, and LC-MS analysis shows two peaks of equal intensity with identical mass spectra. The options provided are: A. Contamination with a precursor - This would likely show a different peak or pattern in the NMR or MS. B. Mixture of diastereoisomers - Diastereoisomers would have different physical properties and would show distinct peaks in NMR and MS. C. Double coupling during an amide-bond forming reaction - This would typically result in more complex splitting patterns due to the coupling constants involved. D. Mixture of enantiomers - Enantiomers would have identical physical properties and would show the same peaks in NMR and MS.\texttt{</reason>}\allowbreak\texttt{<answer>}\allowbreak B. The crude compound exists as a mixture of diastereoisomers \allowbreak\texttt{</\allowbreak answer>}}
\\
\bottomrule
\end{tabular}
\caption{\textbf{Illustration of generated responses under LONG mode} on GPQA, including zero-shot, TTRL and \our. The base model is Qwen3-4B-Base.}
\label{tab:complete_qual_examples_reasoning_gpqa}
\end{table*}

\begin{table*}[t]
\centering

\footnotesize
\setlength{\tabcolsep}{6pt}
\renewcommand{\arraystretch}{1.2}

\begin{tabular}
{@{}p{0.14\textwidth}p{0.82\textwidth}@{}}
\toprule
& \textbf{MATH-500}\\
\midrule

\textbf{Question} &
\parbox[t]{\linewidth}{
{\textbf{Question:} What is the least positive integer multiple of 30 that can be written with only the digits 0 and 2?}
}
\\
\midrule

\textbf{GT} &
\parbox[t]{\linewidth}{2220}
\\
\midrule

\textbf{Zero-shot's outputs} &
\parbox[t]{\linewidth}{To find the least positive integer multiple of 30 that can be written with only the digits 0 and 2, we need to consider the properties of multiples of 30. A number is a multiple of 30 if and only if it is divisible by both 2 and 15. This means the number must end in 0 (to be divisible by 2) and the sum of its digits must be divisible by 3 (to be divisible by 15). Given that the number can only contain the digits 0 and 2, let's start by checking the smallest possible numbers: 1. **Single-digit numbers**: None of the single-digit numbers (0, 2) are multiples of 30. 2. **Two-digit numbers**: The only two-digit number that ends in 0 is 20, but 20 is not divisible by 15 (since $2 + 0 = 2$ is not divisible by 3). 3. **Three-digit numbers**: We need to find a number ending in 0 where the sum of the digits is divisible by 3. Let's check:  - 200: ($2 + 0 + 0 = 2$) (not divisible by 3)  - 220: ($2 + 2 + 0 = 4$) (not divisible by 3)  - 202: ($2 + 0 + 2 = 4$) (not divisible by 3)  - 222: ($2 + 2 + 2 = 6$) (divisible by 3) Since 222 is divisible by 3 and ends in 0, it is a multiple of 30. Therefore, the least positive integer multiple of 30 that can be written with only the digits 0 and 2 is 222. \texttt{<reason>}\allowbreak The number must end in 0 to be divisible by 2 and the sum of its digits must be divisible by 3 to be divisible by 15. Checking the smallest possible numbers, we find that 222 meets both criteria.\texttt{</reason>}\allowbreak \allowbreak\texttt{<answer>}\allowbreak 222\allowbreak\texttt{</\allowbreak answer>}}
\\
\midrule

\textbf{TTRL's outputs} &
\parbox[t]{\linewidth}{\texttt{<reason>}\allowbreak To find the least positive integer multiple of 30 that can be written with only the digits 0 and 2, we need to consider the properties of multiples of 30. A number is a multiple of 30 if and only if it is a multiple of both 3 and 10.  1. **Multiple of 10**: A number is a multiple of 10 if its last digit is 0. Therefore, the number we are looking for must end in 0. 2. **Multiple of 3**: A number is a multiple of 3 if the sum of its digits is a multiple of 3. Since the only digits we can use are 0 and 2, we need to find the smallest number ending in 0 where the sum of the digits is a multiple of 3. Let's start constructing such a number: - The smallest number ending in 0 is 20 (2 + 0 = 2, not a multiple of 3). - The next candidate is 220 (2 + 2 + 0 = 4, not a multiple of 3). - The next candidate is 200 (2 + 0 + 0 = 2, not a multiple of 3). - The next candidate is 2200 (2 + 2 + 0 + 0 = 4, not a multiple of 3). - The next candidate is 2000 (2 + 0 + 0 + 0 = 2, not a multiple of 3). - The next candidate is 2220 (2 + 2 + 2 + 0 = 6, which is a multiple of 3). Thus, 2220 is the smallest number that meets both criteria: it ends in 0 and the sum of its digits is a multiple of 3. Therefore, the least positive integer multiple of 30 that can be written with only the digits 0 and 2 is 2220.\texttt{</reason>}\allowbreak\texttt{<answer>}\allowbreak 2220\allowbreak\texttt{</\allowbreak answer>}
} 
\\
\midrule

\textbf{\our{} 's outputs} &
\parbox[t]{\linewidth}{\texttt{<reason>}\allowbreak To find the least positive integer multiple of 30 that can be written with only the digits 0 and 2, we need to consider the properties of multiples of 30. A number is a multiple of 30 if and only if it is a multiple of both 3 and 10.  1. **Multiple of 10**: A number is a multiple of 10 if its last digit is 0. Therefore, the number we are looking for must end in 0. 2. **Multiple of 3**: A number is a multiple of 3 if the sum of its digits is a multiple of 3. Since the only digits we can use are 0 and 2, the sum of the digits must be a multiple of 3. Given that the number must end in 0, the possible forms are 2220, 220, 20, etc. We need to check the smallest ones first: - **20**: Sum of digits = 2 + 0 = 2 (not a multiple of 3) - **220**: Sum of digits = 2 + 2 + 0 = 4 (not a multiple of 3) - **2220**: Sum of digits = 2 + 2 + 2 + 0 = 6 (a multiple of 3) Since 2220 is the smallest number that meets both criteria (ends in 0 and the sum of its digits is a multiple of 3), it is the least positive integer multiple of 30 that can be written with only the digits 0 and 2.\texttt{</reason>}\allowbreak\texttt{<answer>}\allowbreak 2220\allowbreak\texttt{</\allowbreak answer>}}
\\
\bottomrule
\end{tabular}
\caption{\textbf{Illustration of generated responses under LONG mode} on MATH-500, including zero-shot, TTRL and \our. The base model is Qwen3-4B-Base.}
\label{tab:complete_qual_examples_reasoning_math500}
\end{table*}

\begin{table*}[t]
\centering

\footnotesize
\setlength{\tabcolsep}{6pt}
\renewcommand{\arraystretch}{1.2}

\begin{tabular}
{@{}p{0.14\textwidth}p{0.82\textwidth}@{}}
\toprule
& \textbf{GSM8K}\\
\midrule

\textbf{Question} &
\parbox[t]{\linewidth}{
{\textbf{Question:} I have 10 liters of orange drink that are two-thirds water and I wish to add it to 15 liters of pineapple drink that is three-fifths water. But as I pour it, I spill one liter of the orange drink. How much water is in the remaining 24 liters?}
}
\\
\midrule

\textbf{GT} &
\parbox[t]{\linewidth}{15}
\\
\midrule

\textbf{Zero-shot's outputs} &
\parbox[t]{\linewidth}{You are a helpful assistant. \texttt{<reason>}\allowbreak First, let's determine the amount of water in the original 10 liters of orange drink. Since it is two-thirds water, the amount of water in the orange drink is (2/3) * 10 = 6.67 liters. Next, we need to account for the spill. Since one liter of the orange drink is spilled, we subtract that from the total amount of water in the orange drink: 6.67 - 1 = 5.67 liters. Now, let's calculate the amount of water in the 15 liters of pineapple drink. Since it is three-fifths water, the amount of water in the pineapple drink is (3/5) * 15 = 9 liters. Finally, we add the remaining water from the orange drink and the water from the pineapple drink to find the total amount of water in the remaining 24 liters: 5.67 + 9 = 14.67 liters. Therefore, there are 14.67 liters of water in the remaining 24 liters.\texttt{</reason>}\allowbreak \allowbreak\texttt{<answer>}\allowbreak 14.67\allowbreak\texttt{</\allowbreak answer>}}
\\
\midrule

\textbf{TTRL's outputs} &
\parbox[t]{\linewidth}{\texttt{<reason>}\allowbreak First, let's determine the amount of water in the original 10 liters of orange drink. Since it is two-thirds water, the amount of water in the orange drink is:$\text{Water in orange drink} = \frac{2}{3} \times 10 = \frac{20}{3} \approx 6.67 \text{ liters}$. Next, let's determine the amount of water in the 15 liters of pineapple drink. Since it is three-fifths water, the amount of water in the pineapple drink is: $\text{Water in pineapple drink} = \frac{3}{5} \times 15 = 9 \text{liters}$. Since you spill 1 liter of the orange drink, the amount of orange drink left is: $\text{Remaining orange drink} = 10 - 1 = 9 \text{liters}$. Now, let's calculate the amount of water in the remaining 9 liters of orange drink: $\text{Water in remaining orange drink} = \frac{2}{3} \times 9 = 6 \text{liters}$. Finally, the total amount of water in the combined mixture is the sum of the water from the remaining orange drink and the pineapple drink: $\text{Total water} = 6 + 9 = 15 \text{ liters}$. So, the total amount of water in the remaining 24 liters is 15 liters.\texttt{</reason>}\allowbreak\texttt{<answer>}\allowbreak 15\allowbreak\texttt{</\allowbreak answer>}
} 
\\
\midrule

\textbf{\our{} 's outputs} &
\parbox[t]{\linewidth}{\texttt{<reason>}\allowbreak First, let's determine the amount of water in the original 10 liters of orange drink. Since the orange drink is two-thirds water, the amount of water in it is:  $\text{Water in orange drink} = \frac{2}{3} \times 10 = \frac{20}{3} \approx 6.67 \text{liters}$. Next, let's determine the amount of water in the 15 liters of pineapple drink. Since the pineapple drink is three-fifths water, the amount of water in it is: $\text{Water in pineapple drink} = \frac{3}{5} \times 15 = 9 \text{liters}$. Since you spill 1 liter of the orange drink, the amount of orange drink left is: $\text{Remaining orange drink} = 10 - 1 = 9 \text{liters}$. Now, let's calculate the amount of water in the remaining 9 liters of orange drink: $\text{Water in remaining orange drink} = \frac{2}{3} \times 9 = 6 \text{liters}$. The total amount of water in the combined mixture is the sum of the water in the remaining orange drink and the water in the pineapple drink: $\text{Total water} = 6 + 9 = 15 \text{liters}$. The total volume of the mixture after spilling is: $\text{Total volume} = 9 + 15 = 24 \text{liters}$. Therefore, the amount of water in the remaining 24 liters is: $\text{Water in the remaining mixture} = 15 \text{liters} $.\texttt{</reason>}\allowbreak\texttt{<answer>}\allowbreak 15\allowbreak\texttt{</\allowbreak answer>}}
\\
\bottomrule
\end{tabular}
\caption{\textbf{Illustration of generated responses under LONG mode} on GSM8K, including zero-shot, TTRL and \our. The base model is Qwen3-4B-Base.}
\label{tab:complete_qual_examples_reasoning_gsm8k}
\end{table*}

\end{document}